\documentclass[11pt]{article}

\usepackage{algorithm}
\usepackage{algorithmic}
\usepackage{agenticlearning-xelatex}
\usepackage{tabularx}

\makeatletter
\newcommand{\blfootnote}[1]{%
  \begingroup
  \renewcommand\@makefntext[1]{##1}
  \footnotetext{#1}%
  \endgroup
}
\makeatother

\newcommand{\model}[1]{\texttt{#1}}

\newcommand{\appref}[1]{Appendix~\ref{#1}}

\begin{document}

\shorttitle{You cannot photograph the same street twice}
\shortauthor{Kaizhen Tan}

\title{You Cannot Photograph the Same Street Twice:\\Reliability Limits in
Vision--Language Measurement of Urban Change}
\author{
  Kaizhen Tan\\[0.12cm]
  \normalsize New York University
}
\date{}
\maketitle

\section*{Highlights}
\begin{itemize}\itemsep1pt \parskip0pt
\item Re-photographing unchanged streets shifts scores by 2/3 of the gap between streets.
\item Half of the shift arises in the scoring; the rest is tied to the photograph.
\item Redeveloped streets score wealthier and better maintained, with less greenery.
\item Streets photographed in the same capture campaign share part of their error.
\item Perception change is reliable for groups of streets, not for single sample points.
\end{itemize}

\begin{abstract}
Vision--language models (VLMs) are increasingly used to measure urban change
from repeated street-level imagery, and the results are often mapped for
individual sample points.
Such maps assume that a perception score changes only if the street does.
We test this assumption with repeated Google Street View captures of unchanged
streets in five US cities and with controlled experiments that hold the
photograph, scene or camera fixed. Re-photographing an unchanged street
shifts its perception score by two-thirds of the average difference between
two streets in the same city. Much of this shift arises in the scoring and
averages out over question orders; the rest comes from the photographs and is
not predicted by image statistics of weather and light. A single image measures
a place moderately well, but the change at one location barely separates
redeveloped from unchanged streets. All models tested score
degraded images as worse-looking streets, and in crowdsourced imagery camera
differences cause false reports of physical change unless both images are
rendered through a common virtual camera. After aggregation, redeveloped streets are
scored as wealthier and better maintained, with more enclosure and less
greenery. Streets photographed in the same capture campaign share part of
their error, which averaging within a district does not remove. For urban
research and planning, VLM
perception scores can support comparisons between groups of streets, such as
redeveloped and unchanged streets photographed in the same campaigns, but not
the identification of individual streets that improved or declined.

\medskip
\noindent\textbf{Keywords:} Street view imagery; Vision--language models;
Urban perception; Urban change; Measurement reliability; Repeat photography
\end{abstract}

\section{Introduction}
\label{sec:intro}

Cities are continuously rebuilt, and how redevelopment alters the perceived
quality of streets is a recurring question in urban research and planning.
Street-level imagery has become a standard data source for studying this
question, and reviews of the field already cover several hundred applications
\citep{biljecki2021street,ito2024understanding}. A growing share of this work
is longitudinal. Two images of one location, captured years apart, are compared
to establish what changed on the ground and how the change affected the way the
street is perceived \citep{naik2017computer,stalder2024self,liu2025physical}.
Such comparisons have been used to test theories of urban change
\citep{naik2017computer}, to detect gentrification at individual properties
\citep{ilic2019deep} and to trace it against earlier field surveys
\citep{hwang2014divergent}. The results are commonly mapped with a change in
perception for each sample point, from which clusters of improvement or decline
are identified \citep{liu2025physical}.

Vision--language models (VLMs) have lowered the cost of this design. A single
prompt returns a judgement of whether and how a street has changed, together
with a set of perception scores, and no task-specific training data are
required. Benchmarks now evaluate VLMs on both tasks
\citep{mushkani2026benchmarks,he2026urbanfeel}.

In such a design the model and its prompt act as a measurement instrument, and
the comparison of two captures assumes that a score changes only when the
street changes. Existing evaluations assess validity instead: the agreement of
VLM perception scores with human ratings of single images
\citep{mushkani2026benchmarks} and the accuracy of their answers to questions
about change \citep{he2026urbanfeel}. A longitudinal comparison also depends on
reliability, that is, on whether a score is reproduced when the same scene is
photographed again. Two captures of one location differ in weather, season,
traffic and camera even when nothing has been built or removed; we call these
differences acquisition conditions. They shift indicators derived from
street-level imagery, including perception scores from models trained on human
ratings \citep{zhao2025seasonal,ki2026weather}, but for a prompted VLM the size
of the resulting variation, and its relation to the differences between places
that a map of perception change is meant to show, have not been measured.

\begin{figure}[!tb]
  \centering
  \includegraphics[width=\linewidth]{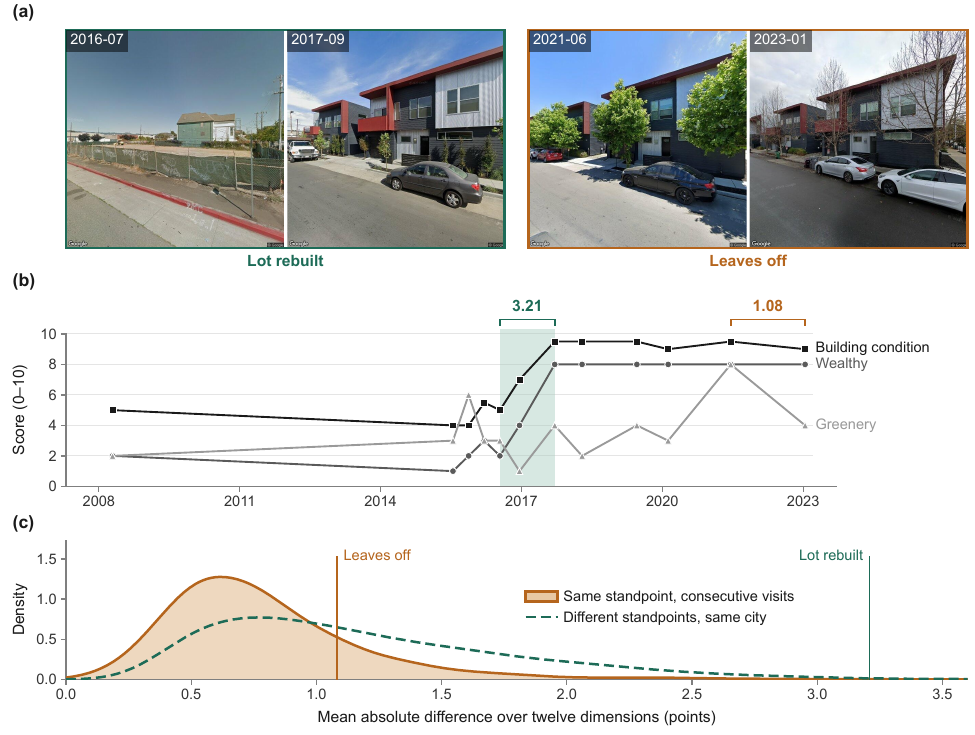}
  \caption{One Oakland standpoint photographed at twelve Street View visits
  from 2008 to 2023 (heading $90^{\circ}$). (a) The lot before and after eight
  townhouses were built (teal; labelled redevelopment) and the same street in
  leaf and then bare (amber; no labelled change). (b) Scores from
  \model{gemini-2.5-flash} at each visit on three of the twelve dimensions;
  brackets give the mean absolute difference over all twelve dimensions for the
  two pairs in (a), and the band spans the labelled redevelopment. (c)
  Densities of that quantity for the 4,648 consecutive-visit frame pairs at
  control standpoints and for all 60,220 pairs of different standpoints in the
  same city (standpoint means). Imagery \textcopyright{} Google.}
  \label{fig:example}
\end{figure}

Figure~\ref{fig:example} illustrates the problem with one location from the
panel used in this study, which we call a standpoint. A vacant lot in Oakland
was photographed by the Street View car twelve times between 2008 and 2023, and
eight townhouses were built on it in 2017. The model follows this
redevelopment: the street is scored as wealthier and better maintained once the
townhouses are complete, and the scores stay at that level for six years.
Between June 2021 and January 2023 nothing was built or removed:
the street trees lost their leaves and the day was overcast. Asked with the
strictest change-detection prompt of this study, the model reports no physical
change between these two visits, yet its perception scores differ by 1.08
points, about 90\% of the average difference between two standpoints in the
same city. The scoring procedure does not mark this reading as unreliable, and
on a map it would appear as change.

Whether this example is typical determines whether reporting at the level of
single points is justified. This study therefore measures how repeatable VLM
perception scores are at locations that did not change. The main evidence is a
panel of
Google Street View standpoints from the CityPulse dataset
\citep{huang2024citypulse}, in which annotators inspected every capture and
marked the points at which substantial redevelopment occurred. Standpoints that
were never marked provide repeated photographs of unchanged streets. Three
controlled experiments on fixed images complement the panel, each isolating one
source of variation. Four research questions (RQs) are addressed.

\begin{description}
  \item[RQ1.] How much does a perception score change when an unchanged street
  is photographed again, compared with the difference between streets and with
  the variation caused by the model and the prompt?
  \item[RQ2.] Can this change be attributed to acquisition conditions and
  corrected accordingly?
  \item[RQ3.] What additional error arises in crowdsourced imagery, where the
  camera differs between captures?
  \item[RQ4.] At what level of aggregation does the measurement become
  reliable, and which signal of redevelopment does it recover at that level?
\end{description}

The study contributes to street-view-based urban research in three respects.
First, it measures the repeatability of a score, as distinct from its agreement
with human ratings, on the scale of the differences between places. Second, it
separates the variation that repeated scoring removes from the variation that
remains tied to the photograph. Third, it converts the measured
variation into the number of locations and the length of street needed to
detect an effect of a given size, and it shows that locations photographed in
the same capture campaign share part of their error. This gives analysts and
planners a basis for choosing the spatial unit at which change in perception
is reported and the unchanged streets against which it is compared.

\section{Related work}
\label{sec:related}

\paragraph{Longitudinal measurement from street-level imagery.}
\citet{naik2017computer} established the template for this line of work. They
derived a score of physical change from paired Google Street View images of
2007 and 2014 in five US cities and tested theories of urban change against it.
\citet{stalder2024self} learned embeddings that are insensitive to lighting and
season but sensitive to structural change and mapped change in London.
\citet{liu2025physical} combined a change-detection network with a perception
model trained on Place Pulse 2.0 to study New York City and Memphis from 2007
to 2023, and reported hot-spot and cold-spot clusters of perception change at
300\,m sample points. It is the closest antecedent of this study and reports
results at the point level, the scale examined here.

\paragraph{Acquisition conditions as a source of bias.}
\citet{curtis2013using} found that the capture date of Street View imagery
changes along a street, so that adjacent images can
differ by months and in places by years. \citet{zhao2025seasonal} quantified
seasonal bias in indicators from a semantic segmentation model across forty
cities and showed that it changes a downstream clustering of urban function.
\citet{ki2026weather} used repeat Google Street View imagery and perception
models trained on Place Pulse 2.0 to show that the weather at the moment of
capture shifts perception scores, by different amounts in different cities.
Both studies rely on models trained for a specific task; we ask the same
question for a prompted VLM. \citet{alpherts2025artifacts} and
\citet{fan2025coverage} addressed a
different issue, namely the bias in where crowdsourced imagery exists and how
densely it samples the street.

\paragraph{Reliability of perception measurement.}
Most of this literature relies on human ratings from Place Pulse
\citep{salesses2013collaborative} and its successor Place Pulse 2.0
\citep{dubey2016deep}, and VLMs are now used as a substitute for such ratings.
\citet{mushkani2026benchmarks} evaluated seven VLMs against 100 Montreal scenes
annotated on thirty dimensions by twelve participants, and argued that
benchmarks should report inter-annotator reliability alongside model alignment.
\citet{he2026urbanfeel} benchmarked twenty models on temporal change questions
over multi-temporal imagery from eleven cities and measured answer accuracy.
Both studies assess agreement with a human reference.
Reliability under repeated observation has a longer history in observational
studies of neighbourhoods. \citet{raudenbush1999ecometrics} showed that the
reliability of a neighbourhood measure aggregated from systematic observations
depends on how many observations enter it. Virtual audits based on Street View
agree less well with field audits for
small features and for features that change within a week, such as litter
\citep{rundle2011using}, and their inter-rater reliability has been assessed
\citep{mooney2014validity}. This perspective has not yet been applied to
perception scores obtained from VLMs.

\paragraph{Pairing and viewpoint.}
\citet{torkko2026pairwise} pairs crowdsourced images by feature matches and
semantic-mask agreement, which controls the viewpoint but not the differences
in camera geometry between contributors, whose effect on VLM outputs has not
been measured.

Three gaps follow from this literature. VLM perception scores have been
evaluated against human ratings of single images, but not for their
repeatability across photographs of the same place. Acquisition bias has been
quantified for task-specific models, but for prompted VLMs neither its size
relative to the differences between places nor its predictability from image
properties is known. Longitudinal studies report change for individual sample
points without an estimate of how many observations such a report requires.

\section{Data and methods}
\label{sec:data}

\subsection{Research design}
\label{sec:design}

\begin{figure}[!tbp]
  \centering
  \includegraphics[width=\linewidth]{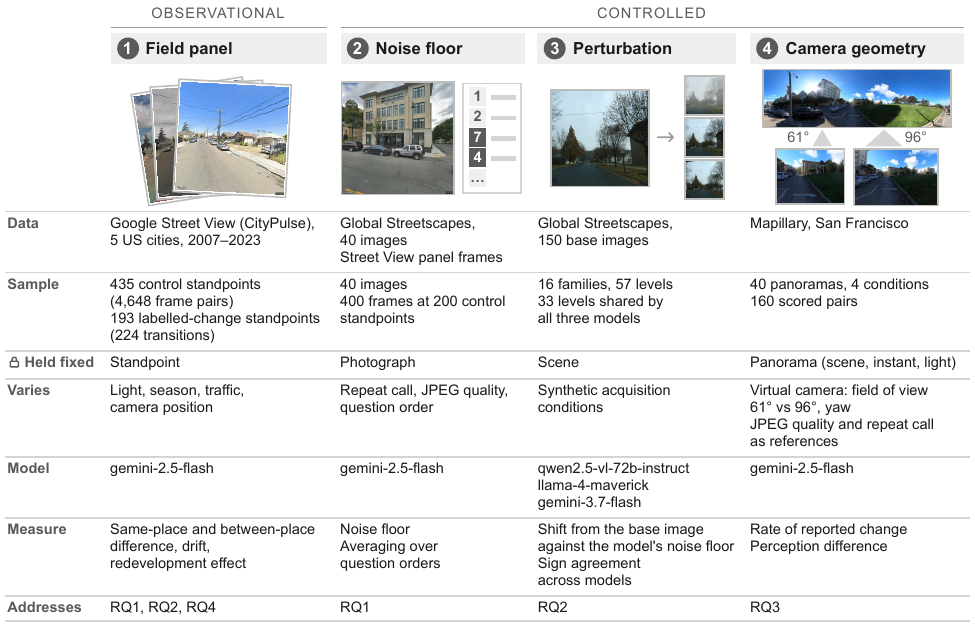}
  \caption{Design of the four experiments, in the order in which they are
  reported; the perturbation strengths shown in (3) are illustrative. Street
  View imagery \textcopyright{} Google; Mapillary and KartaView imagery CC BY-SA
  4.0.}
  \label{fig:design}
\end{figure}

The study combines one observational panel with three controlled experiments,
which differ in what is held fixed while the measurement is repeated
(Figure~\ref{fig:design}). The field panel holds only the location fixed, so
that acquisition conditions vary as they do in practice. The noise-floor
experiment holds the photograph fixed, the perturbation experiment the scene,
and the camera-geometry experiment one panorama. The less an experiment holds
fixed, the more sources of variation its differences contain, so together the
four attribute the variation observed in the field to its sources
(\appref{app:design} lists data, models and dates).

\subsection{Longitudinal street-view panel}
\label{sec:panel}

CityPulse \citep{huang2024citypulse} provides 9,751 Google Street View image
records, covering 8,846 distinct panorama identifiers at 757 kerbside positions
in five US cities from 2007 to 2023. Each image was inspected by annotators and
marked as a change point or not, where a change point indicates substantial
redevelopment. CityPulse places its locations at buildings from the Microsoft
building footprint dataset without stating how they were chosen, and two in
five of its series contain a labelled change, so the panel samples redeveloped
and unchanged streets rather than the street network of the five cities.

A \emph{standpoint} is a kerbside position at which CityPulse records a series of
captures, and a \emph{visit} is one dated capture in that series. A
\emph{panorama} is the Street View capture itself, identified by a panorama
identifier, and a \emph{frame} is one image rendered from a panorama at one
heading. A \emph{frame pair} joins two consecutive visits at one standpoint and
one heading, and a \emph{transition} is the same step with the two headings
averaged. A standpoint without any change point in its series is a
\emph{control standpoint}, and a standpoint with one is a
\emph{labelled-change standpoint}.

Requesting a panorama by its identifier returns that specific capture, so no
proximity matching is involved; consecutive captures at one standpoint are a
median of 1.4\,m apart, and 95\% of them are within 5\,m. All identifiers were
checked against the Street View metadata endpoint in August 2026, and 51.2\% of
them still resolved. The lost panoramas are scattered within series, so 88.5\%
of standpoints retain at least two visits, and capture dates are taken from the
same endpoint (\appref{app:design}). Street View cars photograph a city in
campaigns, so standpoints in one city often share their capture months. We
treat the captures of one city in one month as one \emph{campaign}, so that
each transition spans a pair of campaigns. Each surviving
panorama was requested through the Street View Static API at headings
$0^{\circ}$ and $180^{\circ}$, with an output size of 640 by 640 pixels, the
default $90^{\circ}$ field of view and zero pitch. Heading, field of view and
output size are thus fixed by the request and do not vary across visits. The
headings are compass directions, so depending on the orientation of the street
a frame looks along it or at the facades; this mixture adds to the differences
between places and lowers the ratio $R$. Of 9,052 requested frames, 8,992 were scored; one was not returned, and 59
responses could not be parsed.

\begin{figure}[!tbp]
  \centering
  \includegraphics[width=\linewidth]{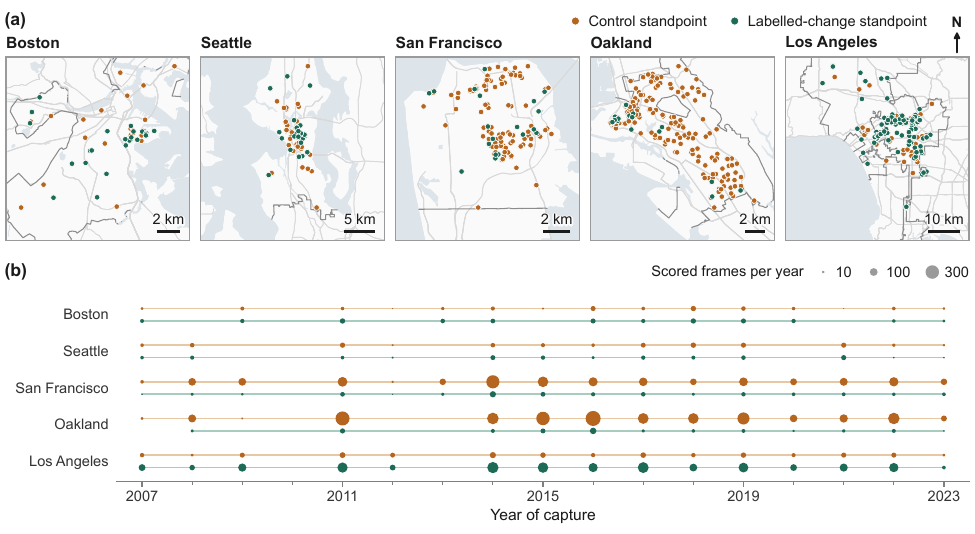}
  \caption{The longitudinal panel. (a) Control (amber) and labelled-change
  (teal) standpoints: 27/27 in Boston, 37/23 in Seattle, 115/20 in San
  Francisco, 202/19 in Oakland and 54/104 in Los Angeles. Maps are north-up,
  each at its own scale; grey lines mark
  city limits on land. Base data: US Census Bureau, TIGER/Line and
  cartographic boundary files, 2023. (b) Years in which each city was
  photographed, control standpoints above and labelled-change standpoints
  below; marker area is proportional to the number of scored frames. The
  median standpoint has 6 to 8 visits, depending on the city.}
  \label{fig:panel}
\end{figure}

The resulting panel contains 435 control standpoints with 4,648 frame pairs
(2,367 transitions) and 193 labelled-change standpoints with 224 transitions
that span a change point. Control and labelled-change standpoints are
distributed unevenly across the five cities (Figure~\ref{fig:panel}), which
Section~\ref{sec:aggregate} takes into account.

\subsection{Imagery for the controlled experiments}
\label{sec:corpus}

Two further sources supply the images for the controlled experiments. Global
Streetscapes \citep{hou2024global} provides 10.0 million street-level images
from 688 cities with per-image acquisition attributes. A manually annotated
subset carries human labels for weather, glare, reflection, quality, platform,
view direction and panorama status. These labels were used to calibrate the
synthetic perturbations and to select the 150 base images of the perturbation
experiment: forward-facing street views labelled as daylight, clear weather and
good quality, without glare or reflection, drawn in near-equal shares per
continent and one city at a time, which spreads them over 109 cities in 50
countries (Figure~\ref{fig:world}). The noise-floor experiment of
Section~\ref{sec:magnitude} uses 40 images from the same corpus, sampled from
its highest-resolution images (median 12 megapixels).

The camera-geometry experiment uses Mapillary imagery over San Francisco;
\appref{app:pairing} describes how the images were listed and paired.

\subsection{Perception scoring}
\label{sec:scoring}

Twelve perception dimensions are scored throughout. Six follow Place Pulse 2.0,
for which human ratings exist: \emph{safe}, \emph{lively}, \emph{beautiful},
\emph{wealthy}, \emph{boring} and \emph{depressing}. The other six are
physically auditable conditions of the kind recorded in street audits
\citep{rundle2011using,mooney2014validity}, with enclosure taken from the urban
design qualities of \citet{ewing2009measuring}: \emph{greenery}, \emph{pavement condition}, \emph{building condition},
\emph{street furniture}, \emph{enclosure} and \emph{visual order}. All twelve
are requested in one prompt. Each dimension is scored from 0 to 10 with anchors
at 0, 5 and 10, and in the field panel and the noise-floor experiment the
model also names the visible evidence for each score. \emph{Boring} and
\emph{depressing} are reverse-coded, and all figures flip them so that a
positive shift always means that the street looks better. For change
detection, the model receives both images of a pair and reports whether a
persistent physical change has occurred and of which type. The prompts are
described in \appref{app:prompts}. All experiments except the perturbation
experiment, which uses three other models (Section~\ref{sec:perturb}), were
scored by \model{gemini-2.5-flash} at temperature zero. The responses
served under an API model identifier can change over time
\citep{chen2024chatgpt}, so every comparison in this study is made within one
scoring batch. The field panel, the noise floor on Global Streetscapes images
and the change-detection evaluation form one batch, and the other experiments
a second; \appref{app:design} gives the access route and the request
parameters.

\subsection{Measures of reliability}
\label{sec:measures}

The repeatability of a score is expressed as a mean absolute difference. Let
$s_k(x)$ denote the score that the model returns for image $x$ on
dimension $k$, with $k=1,\dots,12$. For a set $\mathcal{P}$ of pairs of scored
images, the mean absolute difference on dimension $k$ and its average over
dimensions are
\begin{equation}
  D_k(\mathcal{P}) \;=\; \frac{1}{|\mathcal{P}|}
  \sum_{(x,x')\in\mathcal{P}} \bigl| s_k(x') - s_k(x) \bigr| ,
  \qquad
  D(\mathcal{P}) \;=\; \frac{1}{12}\sum_{k=1}^{12} D_k(\mathcal{P}) .
  \label{eq:mad}
\end{equation}
The quantities compared below differ only in the set $\mathcal{P}$. The
\emph{same-place difference}, $D^{\mathrm{same}}$, is computed on the frame
pairs at control standpoints, that is, on two consecutive photographs of an
unchanged street. Three further conditions hold the photograph fixed: the same
image is scored again with the same request, two copies of the image
re-encoded at JPEG quality 88 and 92 are scored, or the twelve questions are
presented in a different order. We refer to the differences under the last two conditions as
the \emph{noise floor}, since they arise without any change in the photograph.
For each image, the difference is averaged over all distinct pairs of its
scores. The noise floor of \model{gemini-2.5-flash} was measured on 40
Global Streetscapes images in the batch of the field panel, and on 400 panel
frames, the two frames of one control frame pair at each of 200 randomly drawn
control standpoints, in the second batch (Section~\ref{sec:rescore}). For
\model{qwen2.5-vl-72b-instruct} it was measured on the 150 base images of the
perturbation experiment with the prompt used there (\appref{app:design}).
The \emph{between-place difference}, $D^{\mathrm{btw}}$, is computed over all
60,220 pairs of different standpoints in the same city among the 701
standpoints with at least one scored frame, each represented by its mean score
over all of its frames. It equals 1.19 points on
the ten-point scale and
describes the variation that a map of one city is meant to display. Standpoint
means are used because single frames of different standpoints differ by 1.57
points on average, a figure that already contains the frame-level variation
studied here. All other differences are reported as a share of the
between-place difference,
\begin{equation}
  R_k \;=\; \frac{D_k^{\mathrm{same}}}{D_k^{\mathrm{btw}}} ,
  \qquad
  R \;=\; \frac{D^{\mathrm{same}}}{D^{\mathrm{btw}}} .
  \label{eq:share}
\end{equation}
In the terms of observational audits, the test--retest reliability of a single
image as a measure of a place is the intraclass correlation ICC(1) of the
scores of one standpoint view over its visits, computed at control standpoints
with scores centred on their city mean. The perturbation experiment uses a
different unit: shifts are expressed in
standard deviations of the unperturbed scores across its 150 base images,
which we call the between-image standard deviation.

Equation~(\ref{eq:mad}) is evaluated per frame pair, because this is what an
analyst obtains when scoring one image per location per visit. Signed changes
are analysed per transition, because the two headings of a transition are
repeated measurements of the same step. For a transition $\tau$, let
$d_k(\tau)$ be the later score minus the earlier score on dimension $k$,
averaged over the two headings. The \emph{drift} is the mean signed change at
control standpoints. With $\bar d_k(s)$ the mean of $d_k$ over the control
transitions at standpoint $s$, and $\mathcal{S}_0$ the set of control
standpoints,
\begin{equation}
  \delta_k \;=\; \frac{1}{|\mathcal{S}_0|}\sum_{s\in\mathcal{S}_0} \bar d_k(s) ,
  \label{eq:drift}
\end{equation}
so that every standpoint has the same weight. The effect of redevelopment is
the difference between the mean change over the set $\mathcal{T}_1$ of
transitions that span a labelled change point and the mean change over the set
$\mathcal{T}_0$ of control transitions,
\begin{equation}
  \Delta_k \;=\; \frac{1}{|\mathcal{T}_1|}\sum_{\tau\in\mathcal{T}_1} d_k(\tau)
  \;-\; \frac{1}{|\mathcal{T}_0|}\sum_{\tau\in\mathcal{T}_0} d_k(\tau) .
  \label{eq:effect}
\end{equation}
Standard errors of $\delta_k$ and $\Delta_k$ are computed from the variance of
standpoint means, so that the transitions at one standpoint are not treated as
independent observations; Sections~\ref{sec:drift} and~\ref{sec:signal} also
cluster them by pair of campaigns.

Finally, let $\sigma_0$ and $\sigma_1$ be the per-transition standard
deviations of $d_k$ among control and labelled-change transitions, computed
from the variance of standpoint means in the same way as the standard errors;
we use their medians over the twelve dimensions, 0.83 and 1.02 points. The number of
labelled-change transitions needed to detect an effect of size $\Delta$ against
the $n_0 = 2{,}367$ control transitions is
\begin{equation}
  n_1(\Delta) \;=\; \sigma_1^2 \left[
  \left(\frac{\Delta}{z_{1-\alpha/2}+z_{1-\beta}}\right)^{2}
  - \frac{\sigma_0^2}{n_0} \right]^{-1} ,
  \label{eq:power}
\end{equation}
where $z_q$ is the $q$ quantile of the standard normal distribution, the
significance level is $\alpha=0.05$ (two-sided) and the power $1-\beta$ is
0.80.

\section{Repeatability of perception scores at unchanged locations}
\label{sec:repeat}

This section addresses RQ1 and the first part of RQ2.

\subsection{Magnitude of the same-place difference}
\label{sec:magnitude}

\begin{figure}[!tbp]
  \centering
  \includegraphics[width=\linewidth]{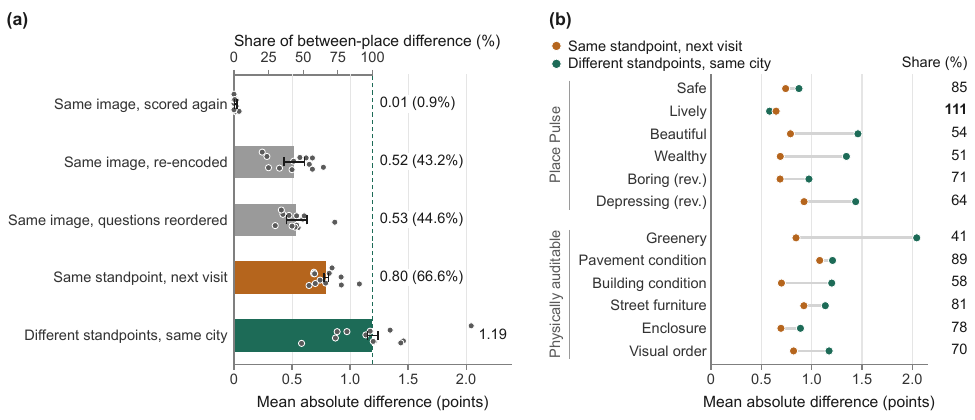}
  \caption{(a) Mean absolute difference in perception score, averaged over the
  twelve dimensions, for the same image scored again, re-encoded at JPEG
  quality 88 and 92, or scored with the questions in two orders (40 Global
  Streetscapes images, scored in the batch of the field panel); for the same
  standpoint at the next visit (4,648 frame pairs); and for different
  standpoints in the same city (standpoint means). Dots show the twelve
  dimensions, and error bars are 95\% cluster-bootstrap intervals. The dashed
  line and the top axis mark the between-place difference. (b) Same-standpoint
  and between-place differences by dimension, with their ratio $R_k$
  (Eq.~\ref{eq:share}) in per cent.}
  \label{fig:repeat}
\end{figure}

Figure~\ref{fig:repeat}a compares the same-place difference with three
conditions that hold the photograph fixed, all as shares of the between-place
difference.

First, in the batch of the field panel the model was close to deterministic
at temperature zero: repeated calls on one image differ by 0.011 points, about
1\% of the between-place difference.

Second, changes to the input that leave the scene untouched move the score
substantially. Re-encoding an image at JPEG quality 88 instead of 92, which gives a
visually indistinguishable file, shifts a score by 0.52 points, 43\% of the
between-place difference. Reordering the twelve questions while leaving the
image untouched shifts it by 0.53 points (45\%).

Third, the same-place difference is the largest of the four. A second
photograph of the same unchanged standpoint differs from the first by 0.80
points, two-thirds of the between-place difference ($R=66.6\%$). The ratio is
large on every dimension
(Figure~\ref{fig:repeat}b): for eleven of the twelve, $R_k$ lies between 41\%
and 89\%, and for \emph{lively} it is 111\%: two photographs of one street
differ more in how lively it looks than two streets in the same city do. The
same-place difference is spread across the panel and
is not concentrated in a few unstable standpoints, since only 7\% to 14\% of
the variance of the pairwise difference lies between standpoints. Averaging the
two headings of a standpoint, which costs one extra request per visit, reduces
the same-place difference by a fifth, to 0.64 points. With the photograph held
fixed, a score already varies by 0.52 to 0.53 points, compared with 0.80 points
for a second photograph.

As a measure of a place at one time, a single image has a test--retest
reliability (ICC) of 0.58 on average over the twelve dimensions, which is
moderate \citep{koo2016guideline}. It is below 0.5 for \emph{pavement
condition} (0.40) and \emph{lively} (0.47) and reaches 0.80 for
\emph{greenery}. The coefficient is computed over all visits of a series, which
at control standpoints span a median of 11 years, so it is a lower bound on the
reliability over short intervals. A change between
two visits is the difference of two such readings, and redevelopment moves a
score by only 0.2 to 0.3 points (Section~\ref{sec:signal}), so a change
measured at one standpoint is far less reliable than a level
(Section~\ref{sec:pointlevel}).

\subsection{Share of the difference that arises in the scoring}
\label{sec:rescore}

\begin{figure}[!tbp]
  \centering
  \includegraphics[width=\linewidth]{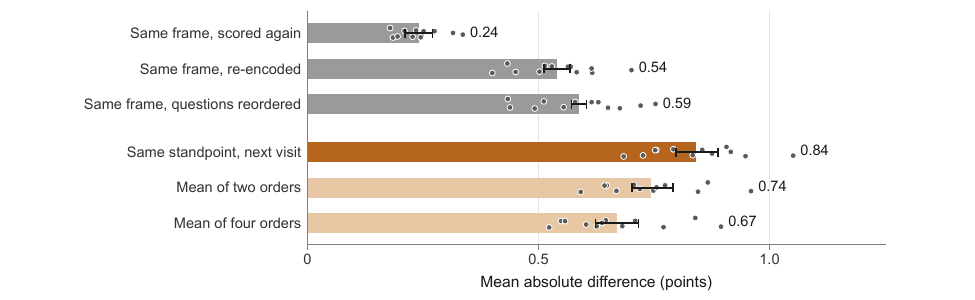}
  \caption{Scoring and photograph as sources of the same-place difference, on
  400 panel frames. Mean absolute difference in perception score between two
  responses to the identical request, to two JPEG encodings and to two of four
  question orders of the same frame, and between the same standpoint at
  consecutive visits, from one question order and averaged over two or four.
  Dots show dimensions; error bars are 95\% bootstrap intervals.}
  \label{fig:rescore}
\end{figure}

How much of the same-place difference arises in the scoring can be measured
only on the panel frames themselves. Both frames of 200 control frame pairs,
one at each of 200 randomly drawn control standpoints, were therefore scored
repeatedly with the field-panel request and with variants that change one
element of it (Figure~\ref{fig:rescore}). In this second batch, repeated
identical requests differ by 0.24 points, against 0.01 in the batch of the
field panel, and the identical request returns scores 0.58 points apart in the
two batches (\appref{app:design}).

Re-encoding a frame shifts a score by 0.54 points and reordering the questions
by 0.59 points, and a second photograph of the same standpoint shifts it by
0.84 points (Figure~\ref{fig:rescore}).
Averaging the scores of four question orders lowers this same-place difference
to 0.67 points, and with more orders it approaches a floor of about 0.6
points, half of the between-place difference. Half of the variance of the
same-place difference thus arises in the scoring and can be averaged away, and
the other half is tied to the content of the two photographs
(\appref{app:design}).

\subsection{Predictability from acquisition conditions}
\label{sec:acquisition}

If the same-place difference were caused by acquisition conditions that can be
measured on the image, it could be modelled and subtracted. To test this, six
statistics were computed for every image: the dark-channel mean
\citep{he2011single}, the whole-frame RMS contrast, the mean saturation, the
mean luminance, the mean absolute Laplacian response and the share of
near-saturated pixels. They measure scattering, contrast, colour, exposure,
sharpness and specularity respectively.

\begin{figure}[!tbp]
  \centering
  \includegraphics[width=\linewidth]{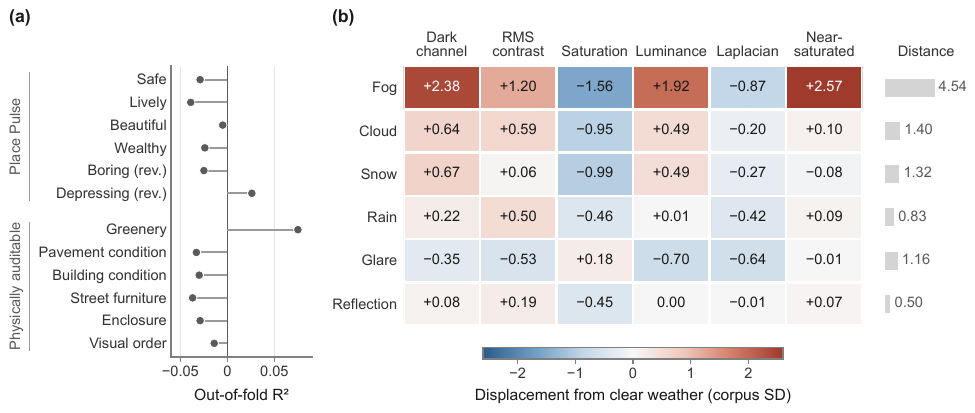}
  \caption{(a) Out-of-fold $R^2$ of the gradient-boosted model of
  Section~\ref{sec:acquisition}, fitted on the 4,648 control frame pairs with
  five folds assigned by standpoint; a negative value means a prediction worse
  than the mean. (b) The
  same six statistics on 4,000 manually annotated Global
  Streetscapes images: displacement of each labelled group from the
  clear-weather mean ($n=598$), in corpus standard deviations. Distance is the
  Euclidean length of the six displacements.}
  \label{fig:acquisition}
\end{figure}

Figure~\ref{fig:acquisition}a relates these statistics to the difference in
score between consecutive visits at the control standpoints. Across all six
statistics and all twelve dimensions, the largest single correlation is 0.19. A
gradient-boosted model was then fitted whose predictors are the levels and the
between-visit differences of the six statistics, the interval between captures
and the seasonal offset, with cross-fitting folds assigned by standpoint. Its
out-of-fold $R^2$ is negative on ten of the twelve dimensions and reaches at
most 0.075, for \emph{greenery}, against an upper bound of about 0.5 set by the
half of the variance that arises in the scoring (Section~\ref{sec:rescore}).
These statistics explain almost none of the same-place difference, so there is
no acquisition component of practical size to subtract.

The same statistics computed on 4,000 manually annotated Global Streetscapes
images show what they can detect (Figure~\ref{fig:acquisition}b). Fog shifts
the dark-channel mean by 2.4 corpus standard deviations, reduces the mean
saturation to 38\% of its clear-weather value and increases the share of
near-saturated pixels eightfold, and cloud, snow, rain and glare each have a
distinct profile. Windscreen reflection is the exception: it is a localised
artefact that whole-frame means average out, and images annotated for it have
the same profile as images without one.

\subsection{Systematic drift}
\label{sec:drift}

Averaging can remove the same-place difference only if it is centred on zero.
We therefore estimate the drift $\delta_k$ of Eq.~(\ref{eq:drift}) on the 2,367
control transitions. The drift is largest for \emph{wealthy} (0.118 points) and
\emph{building condition} (0.110 points), and it is distinguishable from zero
on eleven of the twelve dimensions, also with standard errors clustered by pair
of campaigns. On all eleven it has the same direction: a later photograph
of the same unchanged standpoint is scored as slightly safer, wealthier, better
maintained and more attractive. The exception is \emph{greenery}, with a drift
of $-0.01$ points.

Averaged over all twelve dimensions, with \emph{boring} and \emph{depressing}
reversed, the drift grows by 0.028 points per year of interval between the two
captures (95\% interval 0.015 to 0.041 with standard errors clustered by
standpoint, 0.014 to 0.042 when they are also clustered by pair of campaigns)
and does not change with the calendar year ($-0.004$ points per year,
$p=0.13$). It therefore accumulates at a steady rate with the time
elapsed between captures. Two processes produce such a trend: small physical
change that is not recorded as redevelopment, such as repainting or new
signage, and a gradual improvement of Street View imagery, which would raise
the scores in the way Section~\ref{sec:perturb} describes. At control
standpoints the frames grow sharper by 0.09 standard deviations per year, and
brighter and more contrasted as well. With the between-capture differences of the six statistics in the
regression, the slope falls to 0.018 points per year (0.004 to 0.032), and the
remainder fits unrecorded physical change or aspects of image quality that the
statistics do not capture. The mean absolute same-place difference also rises
with the interval, from 0.75 points under one year to 0.91 points at four to
eight years. Averaging therefore leaves about a tenth of a point of drift.

\subsection{Cities and seasons}
\label{sec:cities}

The same-place difference is smallest in Oakland and Los Angeles (0.74 and
0.79 points) and largest in Boston and Seattle (0.88 and 0.90), with San
Francisco in between (0.83), and $R$ ranges from 62.5\% in Los Angeles to
78.0\% in Boston (Table~\ref{tab:cities}). Matching the season of the two visits
changes little: with the interval held constant, visits in the same season
differ by 0.03 points less (95\% interval 0.00 to 0.06), and on
\emph{greenery} by 0.06 points less. The exception is \emph{greenery} in
Boston and Seattle, which differs by 1.51 points between a leaf-on and a
leaf-off visit and by 0.87 points within one leaf state.

\begin{table}[!tbp]
  \centering
  \caption{Same-place and between-place differences by city, in points, and
  their ratio $R$, with 95\% intervals from resampling the standpoints of each
  city. Same-place differences are computed on control frame pairs,
  between-place differences over pairs of standpoint means in the city.}
  \label{tab:cities}
  \small
  \setlength{\tabcolsep}{5pt}
  \begin{tabular}{@{}lrrlrlr@{}}
    \toprule
    City & \shortstack[r]{Control\\standpoints} & \shortstack[r]{Frame\\pairs}
      & Same-place & \shortstack[r]{Between-\\place} & $R$ (\%)
      & \shortstack[r]{Same-place,\\\emph{greenery}} \\
    \midrule
    Boston        &  27 &   259 & 0.88 [0.82, 0.97] & 1.13 & 78.0 [69.8, 87.9] & 1.02 \\
    Seattle       &  37 &   317 & 0.90 [0.79, 1.06] & 1.21 & 74.4 [64.2, 85.0] & 0.94 \\
    San Francisco & 115 & 1,551 & 0.83 [0.80, 0.86] & 1.13 & 73.6 [67.5, 80.1] & 0.90 \\
    Oakland       & 202 & 2,046 & 0.74 [0.72, 0.76] & 1.18 & 62.7 [58.8, 67.1] & 0.78 \\
    Los Angeles   &  54 &   475 & 0.79 [0.75, 0.85] & 1.27 & 62.5 [56.6, 69.1] & 0.77 \\
    \midrule
    All           & 435 & 4,648 & 0.80 [0.78, 0.82] & 1.19 & 66.6 [63.7, 69.7] & 0.84 \\
    \bottomrule
  \end{tabular}
\end{table}

\subsection{Implication for the reporting scale}
\label{sec:scale}

These results answer RQ1 and the first part of RQ2. A second photograph of an
unchanged street shifts a score by two-thirds of the between-place difference.
About half of the variance of this shift arises in the scoring and can be
averaged away over question orders. The other half is tied to the photographs,
cannot be predicted from whole-image acquisition statistics and is nearly
centred on zero, so at a single location it behaves as random error. A map of
perception change at sample points becomes informative only after aggregation
(Section~\ref{sec:aggregate}).

\section{Acquisition conditions under controlled perturbation}
\label{sec:perturb}

Section~\ref{sec:acquisition} found no predictable acquisition component in
Street View imagery, whose camera is fixed by the request. To answer RQ2 for
conditions that vary more widely, synthetic perturbations were applied to a
fixed set of base images, which holds the scene fixed by construction.

Sixteen families of perturbation were applied to the 150 base images at 57
levels in total, each family reproducing one axis along which the corpus
varies (\appref{app:families}). Atmospheric scattering follows the Koschmieder
model. Its strength and that of rain were fitted to the statistical signature
of the real condition in the annotated corpus (Figure~\ref{fig:acquisition}b),
which gives $\beta=0.95$ for fog, so that $\beta=2.0$ is about twice the
strength of real fog. The response to a perturbation is the mean difference
between the score of a perturbed image and that of its unperturbed base image,
both from the same model.

\begin{figure}[!tbp]
  \centering
  \includegraphics[width=\linewidth]{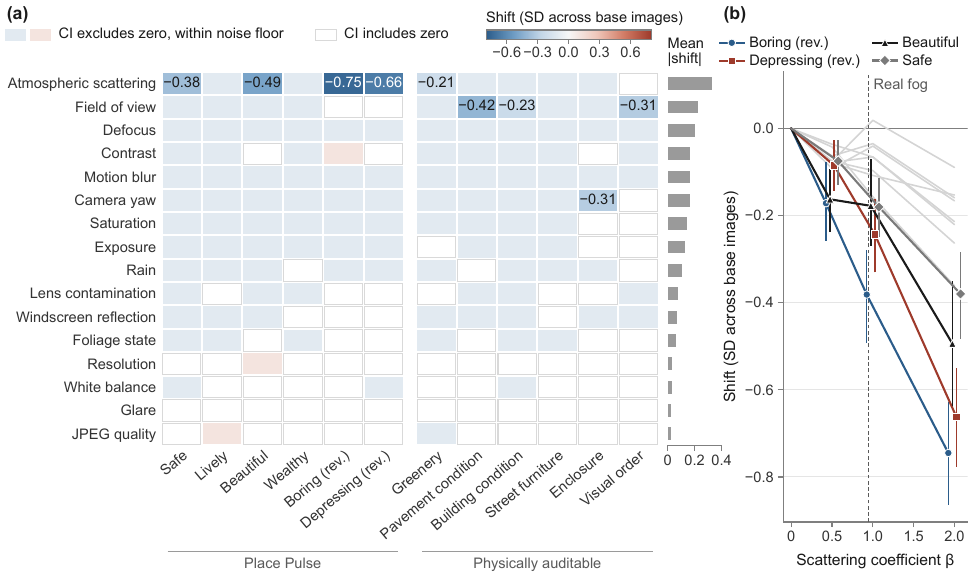}
  \caption{Response of \model{qwen2.5-vl-72b-instruct} to sixteen
  perturbation families, in between-image standard deviations; negative values
  mean that the street looks worse. (a) Shift at the level where each family's
  mean absolute shift (right) is largest. Cells with a value pass the bootstrap test (95\% interval excludes
  zero) and exceed the model's noise floor for reordering on the base images
  (0.24 to 0.64 points, by dimension); pale cells
  pass only the bootstrap test and show its sign; white cells pass neither. (b)
  Shift against the scattering coefficient $\beta$, with 95\% intervals for the
  four largest responses; the dashed line marks the calibrated strength of real
  fog.}
  \label{fig:perturbation}
\end{figure}

Figure~\ref{fig:perturbation}a shows the response of
\model{qwen2.5-vl-72b-instruct}, one of the models used in published
urban-perception benchmarks \citep{mushkani2026benchmarks}, in units of the
between-image standard deviation. Each family is shown at the level where its
mean absolute response is largest, so each row is an upper bound on the effect
of that family. At these levels, 119 of the 192 cells have a bootstrap interval
that excludes zero, 116 of them in the direction of a worse-looking street,
and every family except glare has at least one such cell. Nine cells also exceed the
noise floor of the same model on the same base images
(Section~\ref{sec:measures}), all towards a worse-looking street: five for
atmospheric scattering, three for the narrowest field of view and one for the
strongest camera yaw. All nine lie at the strongest level of their family, and none
of these levels is among the 33 that correspond to conditions found in the
corpus (\appref{app:families}).

The dose--response relationship is clearer than single cells, because model
variability does not grow with perturbation strength whereas a real response
does. For atmospheric scattering, the family with the largest effect, the shift
at $\beta=1.0$, close to real fog, is largest on \emph{boring} and
\emph{depressing} (0.38 and 0.24 between-image standard deviations towards a
worse-looking street), and from $\beta=1.0$ to $\beta=2.0$ it grows on all
twelve dimensions (Figure~\ref{fig:perturbation}b). Acquisition conditions
therefore shift VLM scores systematically. At the strength of real conditions these shifts are smaller
than the noise floor of a single image, but unlike the noise floor they do not
average out when many observations are combined.

Two further models, \model{llama-4-maverick} and
\model{gemini-3.7-flash}, scored the same base images at the 33 levels that
correspond to conditions in the corpus (\appref{app:models}). All three agree
on the sign of the response for ten of the twelve dimensions, and every
response whose 95\% interval excludes zero points towards a worse-looking
street, but which dimensions respond, and how strongly, differs between the
models. A correction estimated with one model would therefore
point the right way under another but would not carry over in magnitude.

\section{Camera geometry in crowdsourced imagery}
\label{sec:camera}

RQ3 concerns crowdsourced imagery, which differs from Street View in two
respects that matter for longitudinal comparison. No panorama identifier links
repeated captures of one place, so pairs have to be identified from metadata,
and the camera varies between contributors.

Both were examined with the Mapillary imagery over San Francisco. Of 300
cross-year candidate pairs drawn at random (\appref{app:pairing}), 32.7\%
lie within 10\,m and $20^{\circ}$ of each other when the position and heading
that Mapillary refines by structure-from-motion are used. Within these pairs
the camera differs: a 2015 dashcam covers $61^{\circ}$ by $48^{\circ}$, a 2024
action camera $96^{\circ}$ by $64^{\circ}$, and an equirectangular panorama the
whole sphere, so two images taken from the same point in the same direction can
frame the street very differently. Both images of a pair were therefore
resampled into one virtual pinhole camera, whose field of view is derived for
each pair from what both sources can cover.

\begin{figure}[!tbp]
  \centering
  \includegraphics[width=\linewidth]{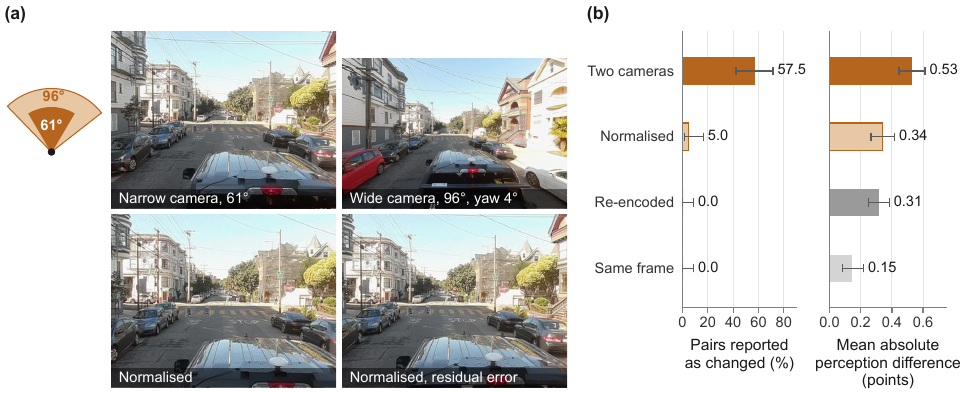}
  \caption{(a) One Mapillary panorama rendered through a narrow
  ($61^{\circ}\times48^{\circ}$) and a wide ($96^{\circ}\times64^{\circ}$,
  yawed $4^{\circ}$) virtual camera (top), and normalised into a common
  $52^{\circ}\times40^{\circ}$ camera (bottom), where the right-hand frame
  keeps $3.1^{\circ}$ of yaw and a 6\% wider field of view as residual error.
  The wedges show the two fields of view in plan. (b) Share of the 40
  panoramas reported as changed, and mean absolute difference in perception
  score, for the two raw cameras, the normalised pair, a normalised frame
  against a copy re-encoded at JPEG quality 88, and the same frame sent twice,
  with Wilson (left) and bootstrap (right) 95\% intervals.}
  \label{fig:camera}
\end{figure}

To isolate the effect of camera geometry, each of 40 panoramas was rendered
twice, once through the narrow camera and once through the wide camera with a
$4^{\circ}$ offset in yaw (Figure~\ref{fig:camera}). Both frames show the same
scene from the same point at the same time, so any change that the model
reports is an artefact. Four conditions were compared: the two raw cameras;
both frames normalised into one virtual camera with a residual misalignment of
$3.1^{\circ}$ in yaw and 6\% in field of view; one normalised frame against a
copy of itself re-encoded at JPEG quality 88 instead of 90; and the same
normalised frame sent twice.

Change detection used the strictest of the three prompts of
\appref{app:threshold}. With the two raw cameras, the model reports
physical change for 23 of the 40 panoramas (57.5\%), most often amenities and
furniture (eight cases), building facades (six), boundary configuration (five)
and road and traffic infrastructure (five). After normalisation the rate falls
to 2 of 40 (5\%), and neither the re-encoded copy nor the repeated frame leads
to a report of change.

Perception scores follow the same pattern. Their mean absolute difference is
0.53 points between the two raw cameras, 0.34 after normalisation, 0.31 between
a frame and its re-encoded copy and 0.15 for the same frame sent twice. The two
cameras add 0.21 points to the difference that re-encoding produces (95\%
interval 0.11 to 0.32), whereas the normalised pair adds 0.02 ($-0.07$ to
0.12). Normalisation thus brings both outputs down to the level of an invisible
change of encoding, and the residual misalignment has no measurable effect.

\section{Reliability after aggregation}
\label{sec:aggregate}

This section addresses RQ4. Averaging removes the random part of the
same-place difference, and a drift shared by two compared groups cancels in
their contrast. The section reports which redevelopment signal remains after
aggregation (Section~\ref{sec:signal}), how well a single transition identifies
redevelopment (Section~\ref{sec:pointlevel}) and how many observations are
needed to resolve an effect of a given size (Section~\ref{sec:samplesize}).

\subsection{Redevelopment signal at the aggregate level}
\label{sec:signal}

\begin{figure}[!tbp]
  \centering
  \includegraphics[width=\linewidth]{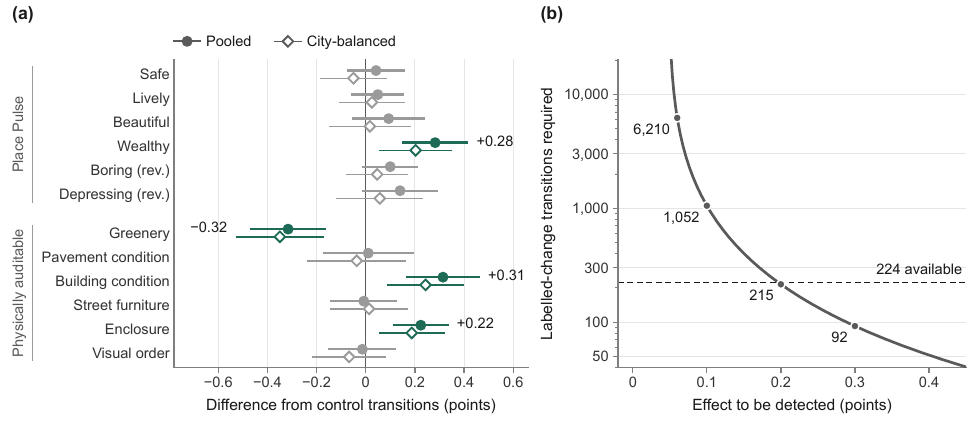}
  \caption{(a) Change in perception score across a labelled change point
  relative to control transitions, in score points with 95\% intervals
  clustered by standpoint; \emph{boring} and \emph{depressing} are reversed.
  Circles show the pooled contrast and diamonds the city-balanced contrast.
  Teal marks the dimensions whose intervals exclude zero. (b) Number of labelled-change transitions required to detect an effect
  of a given size at 80\% power (Eq.~\ref{eq:power}), with the control set held
  at its realised size.}
  \label{fig:aggregate}
\end{figure}

Figure~\ref{fig:aggregate}a shows the effect of redevelopment $\Delta_k$ of
Eq.~(\ref{eq:effect}), estimated from the 224 labelled-change transitions and
the 2,367 control transitions. Four of the twelve dimensions differ between the
two groups, and
together they describe a coherent process. A standpoint that crosses a labelled
change point is scored as wealthier and better maintained, with more enclosure
and less greenery. Redevelopment adds built fabric and removes vegetation, and
the scores register both.

The two groups differ in city composition. Oakland supplies 202 control
standpoints and 19 labelled-change standpoints, whereas Los Angeles supplies 54
and 104, so the pooled contrast partly compares cities. Averaging the
within-city contrasts of the four cities with at least 25 labelled-change
transitions, weighted by those transitions, keeps all four dimensions: \emph{greenery} falls by 0.35 points, \emph{building
condition} rises by 0.24, \emph{wealthy} by 0.20 and \emph{enclosure} by 0.19,
each at $p<0.01$.

Standpoints photographed in the same pair of campaigns share part of their
error (Section~\ref{sec:samplesize}). With standard errors clustered both by
standpoint and by pair of campaigns, all four dimensions remain at $p<0.001$.
A contrast within pairs of campaigns is possible for the 92 labelled-change
transitions that share both capture months with control transitions. It keeps
the sign of all four effects. Those on \emph{greenery} ($-0.31$ points, 95\%
interval $-0.59$ to $-0.03$) and \emph{building condition} (0.29, $-0.01$ to
0.58) keep their size, although the interval of the latter just includes zero,
and those on \emph{wealthy} and \emph{enclosure} fall to 0.16 and 0.07 points
and are not distinguishable from zero. With intervals of this width, only
effects of about 0.3 to 0.4 points or more can be resolved.

\subsection{Change at single locations}
\label{sec:pointlevel}

Taken one transition at a time, the change in score separates the 224
labelled-change transitions from the 2,367 control transitions poorly. The mean absolute change
over the twelve dimensions has an area under the ROC curve of 0.60 (95\%
interval 0.56 to 0.64), and the mean change of the four dimensions that respond
to redevelopment, with \emph{greenery} reversed, reaches 0.65 (0.61 to 0.69).
Among the tenth of transitions with the largest change on this composite, one
in five spans a labelled change point, against one in twelve overall, and
three quarters of the labelled changes lie outside this tenth. A point-level map
of perception change therefore marks mostly streets where nothing was
redeveloped and misses most of those that were. The panel over-samples
redeveloped locations (Section~\ref{sec:panel}), so on a city-wide map the
share of marked streets that were redeveloped would be lower still.

\subsection{Required sample size}
\label{sec:samplesize}

Figure~\ref{fig:aggregate}b converts the measured variances into a required
sample size with Eq.~(\ref{eq:power}). An effect of 0.3 points, close to the
largest effects detected (0.22 to 0.32 points), needs about 92 labelled-change
transitions, and 224 are available. An effect of 0.2 points needs 215 transitions, an
effect of a tenth of a point needs 1,052, and an effect of six hundredths of a
point needs 6,210. An effect of a tenth of a point is no larger than the
drift itself and can be attributed to redevelopment only if both groups share
the same drift. Here they span the same intervals between captures, with
medians of 1.4 and 1.5 years, so the drift cancels in Eq.~(\ref{eq:effect})
and the sample size remains the limiting factor.

For a district, that is, a group of neighbouring locations, the question is
how many locations its mean change needs. With independent errors, the mean
change of a group is resolved to 0.3 points with about 60 locations and to 0.2
points with about 140, which at a sampling interval of 50\,m corresponds to 3
and 7\,km of street. Neighbouring streets are, however, often photographed in
the same campaign: 63\% of the control transitions share both capture months
with another standpoint in their city. Such locations share part of their
error. Their changes correlate at 0.06 (95\% interval 0.04 to 0.08), and at
0.08 within 1\,km of each other, against 0.00 for locations from different
campaigns. A shared month does not imply a shared day, so streets photographed
on the same day probably share more. On \emph{greenery} the correlation is
0.26. It is 0.31 when the two campaigns differ in leaf state and 0.16 when they
do not, since a campaign photographs all of its streets in one leaf state.

For a district photographed in one pair of campaigns, the standard error of its
mean change therefore cannot fall below about 0.2 points, so no change smaller
than about 0.6 points can be resolved, whatever the number of locations. On
\emph{greenery} the floor of the standard error is 0.47 points, larger than the
effect of redevelopment. The change of such a district can be resolved only
against unchanged streets photographed in the same campaigns, where the shared
error cancels. Without such controls, a change of 0.3 points needs locations
from at least four pairs of campaigns, however many locations each pair
contributes.

\section{Discussion}
\label{sec:discussion}

\subsection{Reliability of VLM perception scores at single locations}
\label{sec:disc-reliability}

The central result of this study is that a second photograph of an unchanged
street reproduces two-thirds of the difference between two different streets in
the same city. As a measure of a place at one time, a single image is
moderately reliable. As a measure of change at one location it is not, because
a change is the difference of two readings and redevelopment moves a score by
only 0.2 to 0.3 points, so the change at one location barely separates
redeveloped from unchanged streets. About half of the variance of the
same-place difference arises in the scoring and can be averaged away over
question orders. The other half is tied to the two photographs, acquisition
statistics that clearly separate real weather conditions cannot predict it,
and at the many locations where nothing was redeveloped a point-level map of
perception change displays it.

The result is consistent with earlier evidence that acquisition conditions bias
indicators derived from street-level imagery
\citep{zhao2025seasonal,ki2026weather}, and it adds two elements. The first is
scale: the variation is expressed against the differences between places that
a perception map displays, and on that scale it is of the same order as the
signal. The second is that the variation is not a bias in the usual sense: in
the Street View panel image statistics cannot predict it, and its systematic
part is about a tenth of a point, so it has to be handled by aggregation and
not by correction.

The result also separates reliability from validity. Current benchmarks report
how well VLM scores agree with human ratings of single images
\citep{mushkani2026benchmarks,he2026urbanfeel}; a model can do well on this
and still return different scores for two photographs of one street. As in
observational studies of neighbourhoods \citep{raudenbush1999ecometrics},
reliability grows with the number of observations entering a measure.

\subsection{What redevelopment changes}
\label{sec:disc-urban}

Streets that cross a labelled change point are scored as wealthier and better
maintained, with more enclosure and less greenery, whereas none of the other
Place Pulse dimensions, including safety and liveliness, changes detectably. In
these five cities redevelopment thus registers as new built fabric in good
condition that replaces open lots and vegetation, in line with the loss of
green space reported for densifying cities \citep{haaland2015challenges} and
with the visible signs of reinvestment used to trace gentrification
\citep{hwang2014divergent}, and not as streets that look safer or livelier.
Of the four effects, those on \emph{greenery} and \emph{building condition}
keep their size when the contrast is taken within pairs of campaigns
(Section~\ref{sec:signal}).

The dimensions also differ in how well a photograph records them.
\emph{Greenery}, \emph{beautiful} and \emph{wealthy} are the most repeatable
(ICC 0.80, 0.68 and 0.67), and \emph{pavement condition}, \emph{lively} and
\emph{safe} the least (0.40, 0.47 and 0.57). The first group describes the
fixed fabric of a street, its buildings and vegetation, whereas the second
responds to what the camera catches at the moment of capture: people and
traffic for liveliness and safety, light and wetness on the road surface for
pavement condition. Studies of perceived change can rely more on the first
group, in which \emph{greenery} and \emph{wealthy} also carry the redevelopment
signal. The change of \emph{greenery}, however, depends on the leaf state,
which all streets of a campaign share, so its error is the one most strongly
shared within a campaign (Section~\ref{sec:samplesize}), and its change has to
be compared with unchanged streets photographed in the same campaigns.

\subsection{Implications for longitudinal street-view studies}
\label{sec:implications}

\paragraph{Unit of reporting.}
Change in perception should be reported for groups of locations and not for
single sample points. With the variances measured here, the mean change of a
group is resolved to 0.3 points with about 60 locations and to 0.2 points with
about 140, about 3 and 7\,km of street at a sampling interval of 50\,m,
provided that it is compared with unchanged streets photographed in the same
capture campaigns. Without such controls, the error that a campaign shares
limits what any number of locations can resolve
(Section~\ref{sec:samplesize}). At a
single location, only a change far larger than the same-place difference
stands out from re-photography, as the rebuilt lot in Figure~\ref{fig:example}
does. For planning, VLM scores can therefore support comparisons between
groups of streets, such as redeveloped and unchanged ones, and the evaluation
of programmes that cover many sites, but not the evaluation of a single project
or the identification of individual streets whose perceived quality has
improved or declined.

\paragraph{Study design.}
Two forms of averaging limit the variation before aggregation. Scoring two
headings per location and averaging them lowers the same-place difference by a
fifth at the cost of one extra request, and averaging the scores of four
question orders lowers it by a fifth at the cost of three. Fixing a single
order does not remove the order effect, because the effect of an order differs
from one photograph to the next. Matching the season of the visits helps little outside
the vegetation of seasonal cities. All visits that are compared should be
scored in one batch with one model version, since
the responses under an identifier can change and the acquisition response is
specific to the model, and the prompt, the access route and the request
parameters should be reported with the scores.

\paragraph{Crowdsourced imagery.}
Pairs taken with different cameras should be rendered through a common virtual
camera before they are compared, which in our experiment reduced false reports
of physical change from 57.5\% to 5\% and the perception difference to that of
an invisible re-encoding.

\paragraph{Definition of change.}
The share of pairs in which a VLM reports change depends on how the prompt
defines change: across three prompts, the specificity of change detection
against the CityPulse labels rises from 0.03 to 0.26 at a recall above 0.9
(\appref{app:threshold}). Studies that report rates of detected change should
state the threshold given to the model.

\paragraph{Reproducibility.}
About half of the panorama identifiers published with CityPulse no longer
resolved two and a half years later (Section~\ref{sec:panel}), and the same
request returned different scores in two scoring batches
(Section~\ref{sec:rescore}). Longitudinal studies should record image
identifiers with their retrieval and capture dates, and keep the model
responses together with the request and the serving endpoint.

\subsection{Limitations and future work}
\label{sec:limitations}

The reliability measurement uses Street View imagery, rendered through one
virtual camera at a fixed heading and field of view. Crowdsourced imagery also
differs between contributors in
sensor, lens, mounting height and processing, which the camera-geometry
experiment holds fixed, so its same-place difference has to be measured on such
imagery directly. The panel covers five US cities, and the same measurement in
other regions is left for future work.

Control standpoints are defined by human labels that mark substantial
redevelopment, and \appref{app:threshold} shows that streets also change
in ways that such labels do not record. The part of the drift that the image
statistics leave unexplained (Section~\ref{sec:drift}) and the rise of the
same-place difference with the interval indicate the size of this
contribution. Separating it completely from the
variation caused by re-photography would require labels with an explicit
magnitude threshold.

The field panel was scored with one model. On the same pairs, the second
scoring batch gives a similar same-place difference (0.84 against 0.77
points), and the noise floor of \model{qwen2.5-vl-72b-instruct} is of the same
order (\appref{app:design}), but the same-place difference under other model
families remains to be measured. The decomposition into scoring and photograph
comes from the second batch; in the batch of the field panel the noise floor
was measured on Global Streetscapes images only. The three models in
Section~\ref{sec:perturb} do not include the model
that scored the field panel. Their agreement on direction and disagreement on
size show that the size of a correction must be estimated for the model in use,
which these experiments do not do for \model{gemini-2.5-flash}. Two of the
sixteen perturbation families, glare and reflection, could not be calibrated
against the condition they are named after (\appref{app:families}).
Finally, most base images are 2048-pixel thumbnails with a median of 2.4
megapixels, so the absence of a resolution effect above the noise floor holds
for 0.5 to about 2.4 megapixels.

\section{Conclusion}
\label{sec:conclusion}

Re-photographing an unchanged street shifts a VLM perception score by
two-thirds of the mean difference between two standpoints in the same city. A
single image is a moderately reliable measure of a place, but the change
between two visits at one location barely separates redeveloped from unchanged
streets. About half of the variance of the shift comes from the scoring and can
be averaged away over question orders; the rest is tied to the photographs,
cannot be predicted from image statistics of acquisition conditions and, apart
from a drift of about a tenth of a point, is centred on zero. Degraded images
are scored as worse-looking streets by all three models tested, and camera
geometry in crowdsourced imagery causes false reports of change unless both
images are rendered through a common camera. Aggregated over groups of
streets, redeveloped streets are scored as wealthier and better maintained,
with more enclosure and less greenery. VLM-based measurement of urban change
is therefore suited to comparing groups of streets scored in one batch,
including unchanged streets photographed in the same campaigns, and not to
mapping change at individual sample points.

\section*{Data and code availability}

All primary data are open. Global Streetscapes and Mapillary imagery are
distributed under CC BY-SA, and the CityPulse panorama index was released with
the original paper. Street View imagery is retrieved under the Google Maps
Platform terms and cannot be redistributed, so the release contains panorama
identifiers and derived scores but no images. Of the 8,846 identifiers
published with CityPulse in January 2024, 4,526 still resolved in August 2026.
They are listed with their retrieval date and with the capture date returned
by the metadata endpoint, since this set will continue to shrink.

The analysis code, the sixteen perturbation implementations, the full text of
the five prompt variants, the study-design table and the response cache will be
released on publication. The cache is content-addressed, so every number in the
paper can be regenerated without querying a model again. This is the only way
to reproduce the scores, because the responses served under a model
identifier can change \citep{chen2024chatgpt} and differed between the two
batches of this study (Section~\ref{sec:rescore}).

\section*{Declaration of generative AI and AI-assisted technologies in the
manuscript preparation process}

During the preparation of this work the author(s) used Claude (Anthropic) and
GPT (OpenAI) in order to write and check analysis and plotting code, to edit
the language of the manuscript and to review drafts against the results. After
using these tools, the author(s) reviewed and edited the content as needed and
take(s) full responsibility for the content of the published article. The
vision--language models whose scores are analysed are the object of the study
and are described in Section~\ref{sec:scoring}.

\bibliographystyle{apalike}
\bibliography{main}

\newpage
\appendix
\setcounter{figure}{0}
\setcounter{table}{0}
\renewcommand{\thefigure}{A.\arabic{figure}}
\renewcommand{\thetable}{A.\arabic{table}}
\section{Study design}
\label{app:design}

Table~\ref{tab:design} lists the four experiments with their data source,
sample, unit of inference and model. In the field panel, the Street View
metadata endpoint returns Google's own capture date, which differs from the
date in the published CityPulse index for 2.0\% of the surviving panorama
identifiers. Because the date determines which two images form a consecutive
pair, the endpoint date is used throughout.

All models were accessed through the OpenRouter chat-completions API at
temperature 0, with a JSON schema for the answer and without a seed unless
stated. The paper names each model by its OpenRouter identifier without the
provider prefix, which is \model{google/} for the two Gemini models,
\model{qwen/} for \model{qwen2.5-vl-72b-instruct} and \model{meta-llama/} for
\model{llama-4-maverick}. The output limit is 2,400 tokens for the full prompt
and 600 to 1,400 tokens for the compact one, and 3,000 tokens for
\model{gemini-3.7-flash}, which spends part of its output on reasoning. Images
are sent as JPEG data URLs; panel frames at 640 by 640 pixels and quality 90.
The API returns the identifier and the provider, Google for
\model{gemini-2.5-flash}, but no version string. Every response is stored under
a hash of the full request (identifier, temperature, messages including the
image, schema and seed).

The two scoring batches were two months apart. Without a seed, the
field-panel request for the 400 frames of the decomposition experiment has the
same hash in both batches, so both send the same image files and request body;
the second-batch responses were obtained by adding a seed. The prompt length is
identical for 399 of the 400 frames, the median answer has 646 tokens in both
batches, and neither batch used reasoning tokens. Between the batches, the
identical request differs by 0.58 points on average and is shifted in one
direction on six dimensions, by up to 0.32 points. Within the second batch, the
same request sent twice with the same seed differs by 0.25 points on 100
frames, as much as with different seeds (0.23 points), against 0.67 points at
temperature 1. OpenRouter serves this identifier from several Google endpoints
on Vertex AI and AI Studio; the second batch was served by Vertex AI, and the
endpoint of the first was not recorded.

For the 200 sampled pairs, the first batch gives a same-place difference of
0.77 points, against 0.80 for all control frame pairs. In the second batch, the variance
of the per-pair difference falls linearly with the inverse number of averaged
question orders, from 1.46 for one order to 0.90 for four, towards an intercept
of 0.71, which corresponds to a mean absolute difference of about 0.6 points.
Re-encoding and reordering act as one kind of instability: a re-encoded and a
reordered score of the same frame differ by 0.60 points, as much as two
reordered scores (0.59) and less than the 0.76 points expected if the two
acted independently. Averaging over orders therefore also averages over the
response to invisible changes of the image, and the part that remains is tied
to the content of the photographs.

The noise floor of \model{qwen2.5-vl-72b-instruct}, measured on the 150
perturbation base images with the compact prompt of the perturbation
experiment, is 0.24 points for re-encoding, 0.49 points for reordering and 0.11
points between repeated calls.

\begin{figure}[!htbp]
  \centering
  \includegraphics[width=\linewidth]{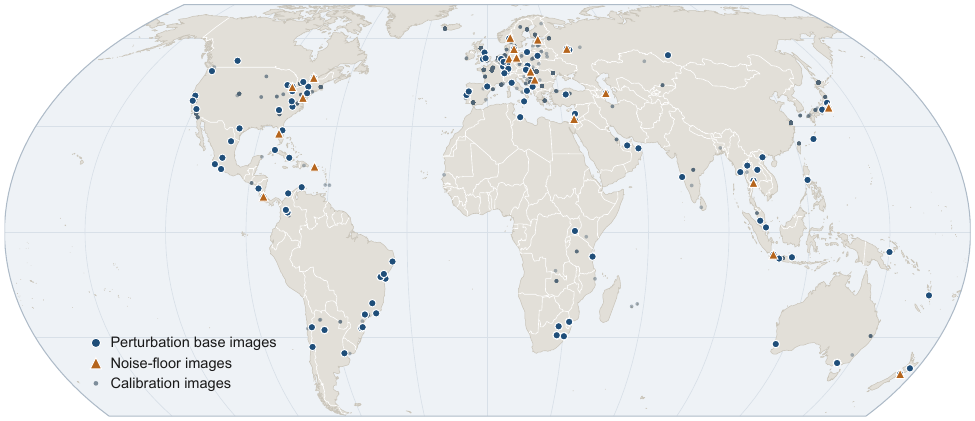}
  \caption{Locations of the images used in the controlled experiments (Equal
  Earth projection): the 150 perturbation base images, the 40 noise-floor
  images and the 4,000 manually annotated Global Streetscapes images used to
  calibrate the perturbations. Land and borders: Natural Earth.}
  \label{fig:world}
\end{figure}

Figure~\ref{fig:world} shows where the images of the controlled experiments
were taken. Within each continent, the base images were drawn one city at a
time in random order, so that the 150 images come from 109 cities in 50
countries. Africa has 21 forward-facing images with clean labels, in seven
cities, and all of them are used; the 25 images from Oceania come from six
cities, and each other continent contributes at most two images per city.

\begin{table}[!htbp]
  \centering
  \caption{The four experiments and the unit at which each is analysed. A frame
  pair joins two consecutive visits at one standpoint and one heading, and a
  transition averages the two headings of the same step. The field panel and
  the noise floor on Global Streetscapes images belong to the first scoring
  batch, the other rows to the second.}
  \label{tab:design}
  \footnotesize
  \setlength{\tabcolsep}{4pt}
  \begin{tabularx}{\linewidth}{@{}l l >{\raggedright\arraybackslash}X >{\raggedright\arraybackslash}X >{\raggedright\arraybackslash}X l@{}}
    \toprule
    Experiment & Source & Scope & Rows analysed & Unit of inference & Model \\
    \midrule
    Field panel     & Street View  & 5 US cities, 2007--2023          & 2,367 + 224 transitions                       & transition, clustered by standpoint & \model{gemini-2.5-flash} \\
    Noise floor     & Streetscapes & 40 images                        & 320 scores (310 parsed)                     & image                               & \model{gemini-2.5-flash} \\
                    & Street View  & 400 frames at 200 control standpoints & 3,200 scores, 400 of them from the first batch & standpoint                   & \model{gemini-2.5-flash} \\
    Perturbation    & Streetscapes & 150 images, 16 families, 57 levels & 9,750 responses per model, 5,100 for \model{gemini-3.7-flash}; 1,194 noise-floor scores & base image & three (Section~\ref{sec:perturb}) \\
    Camera geometry & Mapillary    & 40 panoramas, 4 conditions       & 160 scored pairs                              & panorama                            & \model{gemini-2.5-flash} \\
    \bottomrule
  \end{tabularx}
\end{table}

\section{Perturbation families and levels}
\label{app:families}

Sixteen families are applied at 57 levels in total: atmospheric scattering
($\beta=0.5$, 1.0, 2.0), motion blur (5, 11, 21\,px), defocus (2, 4, 8\,px),
camera yaw (5, 10, $20^{\circ}$), lens contamination (0.3, 0.6, 1.0), rain
(0.3, 0.6, 1.0), glare (0.15, 0.35, 0.6), windscreen reflection (0.2, 0.4,
0.7), foliage state (0.33, 0.66, 1.0), field of view (45, 60, $75^{\circ}$),
resolution (0.5, 1.0, 2.1, 5.1, 8.3, 12.0\,MP), JPEG quality (25, 40, 60, 80,
95), exposure ($\pm0.75$, $\pm1.5$\,EV), white balance (4000, 5000, 8000,
10000\,K), contrast ($\times0.6$, 0.8, 1.25, 1.6) and saturation ($\times0.4$,
0.7, 1.3, 1.8). Thirty-three of these levels correspond to conditions found in
the corpus: resolutions of 1.0, 2.1, 5.1 and 12.0\,MP, within the range of
median image sizes from 2013 to 2023; atmospheric scattering at $\beta=1.0$, close to
the calibrated strength of fog ($\beta=0.95$); rain at 0.6 and 1.0 (calibrated
strength 0.5); foliage states of 0.66 and 1.0; JPEG quality 80 and 60; and the
two mildest levels of each of the other eleven families. These 33 levels are
the ones compared across models in \appref{app:models}. The other 24
levels, 19 of which are stronger than any level of their family in this set,
trace the dose--response.

Every base image is scored once unperturbed and once at each of the 57 levels.
Images are scored with their longer side reduced to 1,024 pixels, which removes
most of the difference between resolution levels, so the unperturbed image and
the six resolution levels are also scored at their native size of up to 2,048
pixels. This gives 65 responses per base image and 9,750 per model. \model{gemini-3.7-flash} scored only the unperturbed image and the 33
levels above, which gives 34 responses per base image and 5,100 in total.

Glare and reflection have no calibrated strength. The glare perturbation adds
a bright source with veiling glare, which raises the mean luminance, whereas
images annotated for glare are darker and lower in contrast than clear-weather
images (Figure~\ref{fig:acquisition}b), so the distance between the two
profiles grows with every increase in strength. Images annotated for
reflection have the same profile as images without one
(Section~\ref{sec:acquisition}). Both families are applied at the three fixed
strengths listed above.

Camera yaw is applied as the exact homography $K R K^{-1}$ for a pinhole image
and by resampling the sphere for an equirectangular one, so the rotation is
exact in both cases and is not approximated by an image shift.

The rain perturbation was revised after calibration. The first version drew
falling streaks, which sharpen the image. Real rainy imagery is softer than
clear imagery, so this version moved the diagnostic statistic in the wrong
direction. The perturbation was therefore rebuilt from the physical
consequences of rain: overcast illumination, a partly specular wet road,
scattering in the air and droplets on the lens.

\section{Direction of the response across models}
\label{app:models}

Direction can be assessed even where the size of a response is below the noise
floor, by averaging the shift over the sixteen families. For each dimension and
model we take this mean and a 95\% interval from resampling the base images.
This was done for three models, \model{qwen2.5-vl-72b-instruct},
\model{llama-4-maverick} and \model{gemini-3.7-flash}, which scored the
identical 150 base images. The comparison is restricted to the 33 perturbation
levels that all three runs share, so that the model is the only factor that
differs.

\begin{figure}[!tbp]
  \centering
  \includegraphics[width=\linewidth]{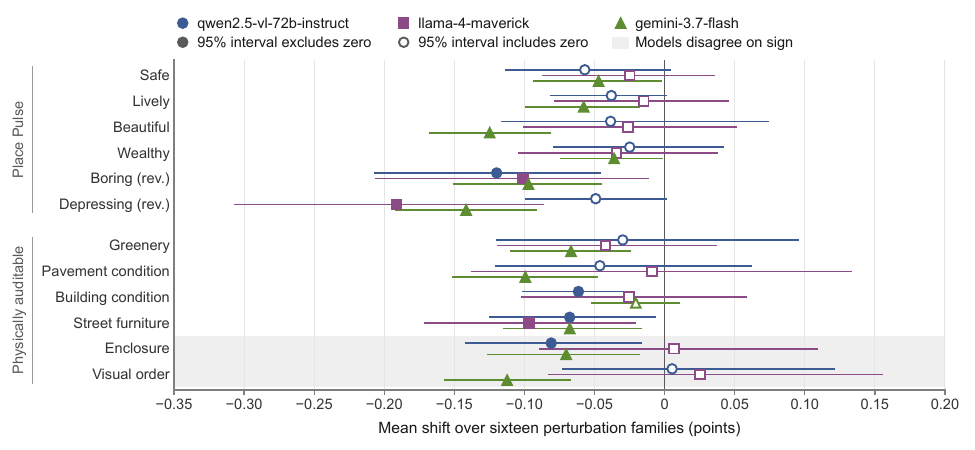}
  \caption{Mean shift over the sixteen perturbation families, by dimension and
  by model, on the same 150 base images at the 33 levels common to the three
  models, in score points (\emph{boring} and \emph{depressing} reversed).
  Lines are 95\% intervals from 2,000 bootstrap resamples of the base images,
  and filled markers mark intervals that exclude zero. Shaded rows mark the
  dimensions on which the models disagree in sign.}
  \label{fig:models}
\end{figure}

Figure~\ref{fig:models} shows the result. The three models agree on the sign
of the response for ten of the twelve dimensions. The interval lies below zero
for 18 of the 36 combinations of model and dimension and above zero for none,
so every response that is distinguishable from zero points towards a
worse-looking street. The models differ in which dimensions respond and by how
much: the interval excludes zero on 11 dimensions for \model{gemini-3.7-flash},
4 for \model{qwen2.5-vl-72b-instruct} and 3 for \model{llama-4-maverick}, and
only for \emph{boring} and \emph{street furniture} under all three; the
correlation of the full family-by-dimension response is 0.51 to 0.65 between
pairs of models. On the two dimensions without a common sign,
\emph{enclosure} and \emph{visual order}, negative responses that exclude zero
meet small positive ones that do not.

The direction of the acquisition response is therefore shared across the three
models, while its size is specific to the model.

\section{Pairing and normalisation of crowdsourced imagery}
\label{app:pairing}

Mapillary images were listed through the vector-tile layer, and per-image
metadata were retrieved through the graph API. Four tiles over San Francisco
returned 441,747 images when retrieved in October 2026. When these are binned by their raw
compass angle into cells of 20\,m by $45^{\circ}$ of heading, 28,155 cells
contain imagery from two or more distinct years. We drew 300 of these cells at
random and paired, within each, a random image from the earliest year with a
random image from the latest year; the pairs spread over the 3.8 by 5.7\,km
covered by the four tiles.

Mapillary provides a position and a heading refined by structure-from-motion,
and a field, \texttt{merge\_cc}, documented as the connected component of
images aligned into a common reconstruction. The refined position
and heading are present on all 600 images and differ from the raw position by a
median of 5.5\,m and from the raw compass heading by a median of
$4.7^{\circ}$. The \texttt{merge\_cc} field cannot serve as a key for
co-located images: only 12 of the 300 pairs (4.0\%) share a value, and the
share is no higher for pairs that lie close together. The pairing criterion was
therefore applied to the refined position and heading alone, and 98 of the 300
pairs (32.7\%) lie within 10\,m and $20^{\circ}$ of each other.

Pairing by keypoint matches, as in the work discussed in
Section~\ref{sec:related}, does not gain from the normalisation of
Section~\ref{sec:camera}. SIFT inliers before
and after normalisation were counted for the 186 pairs within 25\,m and
$35^{\circ}$. The median number of inliers is 15
before normalisation and 13 after, and 27\% of the pairs improve: keypoint
matching is already invariant to the scale and rotation that normalisation
removes, while the resampling loses detail.

For normalisation, both images of a pair are resampled into one virtual
pinhole camera. The yaw applied to each image is the difference between the
target heading and the image's own refined heading, which avoids any dependence
on the orientation convention of the reconstruction. The field of view is
derived for each pair from what both sources can cover. With a fixed
$80^{\circ}$ horizontal field of view instead, a median of 30.0\% of the
earlier frame and 5.1\% of the later frame fall outside the source image,
against 0.4\% and 0.3\% with the derived field of view.

\section{Threshold for reporting change}
\label{app:threshold}

Unrecorded physical change, one of the two sources of drift named in
Section~\ref{sec:drift}, also affects how a rate of detected change should be
read. Table~\ref{tab:change} evaluates paired change detection by
\model{gemini-2.5-flash} against the CityPulse labels. The
evaluation uses four labelled-change standpoints, two in Los Angeles and two in
Oakland, whose series were scored at four headings ($0^{\circ}$, $90^{\circ}$,
$180^{\circ}$ and $270^{\circ}$). Every consecutive pair of visits is
evaluated at each heading, which gives 172 pairs from 43 steps; 24 of the pairs
(6 steps) span a labelled change point. The four headings of one step show the
same street, so the 172 pairs are not independent observations. Three
prompts were applied to the same images with the same model. The first contains
only the task description, the second adds a list of acquisition artefacts to
disregard, and the third further adds a threshold that defines substantial
change (\appref{app:prompts}).

\begin{table}[!tbp]
  \centering
  \caption{Paired change detection by \model{gemini-2.5-flash} against the
  CityPulse labels, which mark substantial redevelopment, on 172
  consecutive-visit pairs. TP, FP, TN and FN are true positives, false
  positives, true negatives and false negatives.}
  \label{tab:change}
  \small
  \setlength{\tabcolsep}{4.5pt}
  \begin{tabular}{@{}lrrrrrrrr@{}}
    \toprule
    Prompt & Reports change & TP & FP & TN & FN & Precision & Recall & Specificity \\
    \midrule
    Task description only                & 97.7\% & 24 & 144 &  4 & 0 & 0.143 & 1.000 & 0.027 \\
    + acquisition artefacts to disregard & 86.0\% & 23 & 125 & 23 & 1 & 0.155 & 0.958 & 0.155 \\
    + threshold for substantial change   & 76.7\% & 22 & 110 & 38 & 2 & 0.167 & 0.917 & 0.257 \\
    \bottomrule
  \end{tabular}
\end{table}

\begin{figure}[!tbp]
  \centering
  \includegraphics[width=\linewidth]{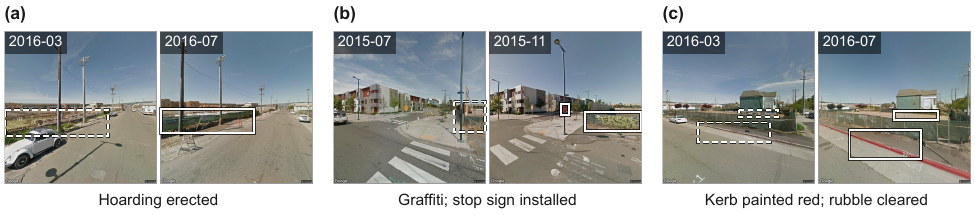}
  \caption{Three false positives of the strictest change-detection prompt at
  the Oakland standpoint of Figure~\ref{fig:example}, four months apart
  each. None spans a labelled change point, and each shows
  a persistent physical change: (a) a construction hoarding erected along the
  fence line; (b) fresh graffiti on the fence and a new stop sign; (c) the kerb
  painted red and rubble cleared from the lot. Pairs (a) and (c) end five
  months before the first change point that CityPulse labels at this
  standpoint. Boxes mark the changed elements (solid) and the same region in
  the earlier image (dashed).}
  \label{fig:unlabelled}
\end{figure}

Specificity rises from 0.027 to 0.257 across the three prompts, while recall
falls only from 1.000 to 0.917. Precision stays near 0.16, because even with the strictest
prompt the model reports change for 76.7\% of the pairs, whereas the reference
label marks 14.0\% of them. Part of this gap is a difference in definition.
The three false positives of the strictest prompt with the shortest interval
between captures (four months) and the smallest acquisition difference, all at
the Oakland standpoint of Figure~\ref{fig:example}, each contain a persistent
physical change that the reference label does not record
(Figure~\ref{fig:unlabelled}). CityPulse labels substantial redevelopment,
whereas the task description asks about any persistent physical change. The
threshold that separates the two definitions is a free parameter of the
measurement and is rarely stated in this literature.

\section{Prompt texts}
\label{app:prompts}

Five prompts are used, all with the same system message. Two score perception
and differ only in the form of the answer. The full version asks for an
evidence clause and a confidence for each dimension and is used for the field
panel and the noise floor; the compact version asks for the scores only and is
used for the perturbation and camera-geometry experiments. Three prompts detect
change. Their common task description defines physical change as a difference
that is material, persistent and within the street corridor, and asks for the
type and valence of each change. The second variant adds the list of
acquisition artefacts below, and the third adds the magnitude threshold below
as well; the camera-geometry experiment uses the third. The two blocks are the
manipulated parts of the prompts behind Table~\ref{tab:change} and are given in
full. The complete prompt texts are released with the code.

\paragraph{The list of acquisition artefacts.}
{\small
\begin{verbatim}
Do NOT report any of the following as physical change. Each of these is an
artefact of how and when the photograph was taken:

- different vehicles parked or moving; different people or cyclists present
- different weather: rain, snow cover, wet or dry road, cloud, fog
- different season: the same trees in leaf versus bare, grass green versus
  brown
- different time of day, sun angle, shadow pattern, street lights on or off
- different image resolution, sharpness, compression, colour rendition,
  exposure
- different camera position or heading, so the same objects are framed
  differently, appear larger or smaller, or fall in or out of view
- different projection (flat perspective versus panoramic) or field of view
- temporary objects: refuse bins out for collection, market stalls, event
  furniture, temporary barriers, construction vehicles passing through
- an object occluded in one image and visible in the other

If an element is present in one image and absent in the other, decide
whether it was removed or merely out of frame or occluded. If you cannot
tell, do not report it.
\end{verbatim}
}

\paragraph{The magnitude threshold.}
{\small
\begin{verbatim}
REPORT ONLY SUBSTANTIAL CHANGE.

A change is substantial if it alters what the street is or how it works,
not merely how it looks on the day. Report it when a structure, surface or
fixed element has been built, demolished, replaced, widened, narrowed or
rebuilt.

Do not report:
- graffiti, tags, posters, stickers or repainting of an existing surface
- kerb or road markings repainted in the same configuration
- construction hoarding, scaffolding, cones or temporary site barriers,
  unless the building behind them has visibly changed
- a plant that has grown, been trimmed, or changed with the season, where
  the planting itself is unchanged
- litter, dumped items, wear, staining, or a surface that is merely
  dirtier
- individual items of street furniture appearing or disappearing

The test to apply: would a planner recording what happened on this street
write this down. If the honest answer is that the street is the same street
with a different surface appearance, report no change.
\end{verbatim}
}

\section{Released tables}
\label{app:tables}

The following tables are released with the code: the per-dimension values
behind Figure~\ref{fig:repeat}, the signed drift by dimension and by capture
era, the decomposition of the pairwise difference into between-standpoint and
within-standpoint variance, the full condition profiles behind
Figure~\ref{fig:acquisition}b, the family-by-level responses behind
Figure~\ref{fig:perturbation}, and the city-by-city contrasts behind
Figure~\ref{fig:aggregate}.

\end{document}